\documentclass[conference]{IEEEtran}
\IEEEoverridecommandlockouts
\usepackage{cite}
\usepackage{amsmath,amssymb,amsfonts}
\usepackage{algorithmic}
\usepackage{graphicx}
\usepackage{textcomp}
\usepackage{xcolor}
\usepackage{array}
\usepackage{float}
\usepackage{placeins}
\usepackage[backref=page,hidelinks]{hyperref}
\usepackage[top=0.78in, bottom=1.05in, left=0.67in, right=0.67in]{geometry}
\def\BibTeX{{\rm B\kern-.05em{\sc i\kern-.025em b}\kern-.08em
    T\kern-.1667em\lower.7ex\hbox{E}\kern-.125emX}}
    \makeatletter
\newcommand{\linebreakand}{%
  \end{@IEEEauthorhalign}
  \hfill\mbox{}\par
  \mbox{}\hfill\begin{@IEEEauthorhalign}
}
\makeatother

\begin{document}

\title{Surformer v2: A Multimodal Classifier for Surface
Understanding from Touch and Vision\\
}

\author{%
\IEEEauthorblockN{Manish Kansana}
\IEEEauthorblockA{\textit{Department of Computer Science and Engineering}\\
\textit{Mississippi State University}\\
Mississippi State, USA\\
mk1684@msstate.edu}
\and
\IEEEauthorblockN{Sindhuja Penchala}
\IEEEauthorblockA{\textit{Department of Computer Science}\\
\textit{The University of Alabama}\\
Tuscaloosa, USA\\
spenchala@crimson.ua.edu}
\linebreakand 
\IEEEauthorblockN{Shahram Rahimi}
\IEEEauthorblockA{\textit{Department of Computer Science}\\
\textit{The University of Alabama}\\
Tuscaloosa, USA\\
srahimi1@ua.edu}
\hspace{7cm}
\and
\IEEEauthorblockN{Noorbakhsh Amiri Golilarz}
\IEEEauthorblockA{\textit{Department of Computer Science}\\
\textit{The University of Alabama}\\
Tuscaloosa, USA\\
noor.amiri@ua.edu}
}

\maketitle

\begin{abstract}
Multimodal surface material classification plays a critical role in advancing tactile perception for robotic manipulation and interaction. In this paper, we present Surformer v2, an enhanced multi-modal classification architecture designed to integrate visual and tactile sensory streams through a late( decision level) fusion mechanism. Building on our earlier Surformer v1 framework \cite{Surformer}, which employed handcrafted feature extraction followed by mid-level fusion architecture with multi-head cross-attention layers, Surformer v2 integrates the feature extraction process within the model itself and shifts to late fusion. The vision branch leverages a CNN-based classifier (Efficient V-Net), while the tactile branch employs an encoder-only transformer model, allowing each modality to extract modality-specific features optimized for classification. Rather than merging feature maps, the model performs decision-level fusion by combining the output logits using a learnable weighted sum, enabling adaptive emphasis on each modality depending on data context and training dynamics. We evaluate Surformer v2 on the Touch and Go dataset \cite{b9}, a multi-modal benchmark comprising surface images and corresponding tactile sensor readings. Our results demonstrate that
Surformer v2 performs well, maintaining competitive inference speed, suitable for real-time robotic applications. These findings underscore the effectiveness of decision-level fusion and transformer-based tactile modeling for enhancing surface understanding in multi-modal robotic perception.
\end{abstract}


\section{Introduction}

Perceiving and understanding surface properties is a crucial capability for autonomous robotic systems operating in real-world environments. While visual perception offers valuable global information, it often falls short in scenarios involving low lighting, occlusion, or visually ambiguous materials. Tactile sensing, on the other hand, provides complementary information such as texture, stiffness, and friction, which are directly relevant to physical interactions. As robotic systems increasingly interact with unstructured environments, leveraging both visual and tactile cues for robust surface understanding has become a central research challenge in multimodal perception.

In recent years, advances in tactile sensors, such as the GelSight family, have enabled high-resolution,image-like tactile data acquisition, opening up new opportunities for multimodal learning. Deep learning and convolutional neural networks (CNNs) based approaches \cite{resnet, densenet, lenet, efficient} can also be applied to both visual and tactile modalities. However, fusing these modalities effectively remains non-trivial due to their distinct characteristics and differing information content. The question is no longer whether to use multi-modal data, but how
best to integrate it for optimal performance and generalization in robotic perception tasks.

Prior to the deep learning models, visual-tactile surface/object understanding in robotics relied on classical machine learning and statistical methods\cite{b1} \cite{b2}. Researchers used Support Vector Machines (SVMs), Regularized Least Square (RLS), Regularized Extreme Learning Machine (RELM) for material and texture classification \cite{b3}. For Tactile data, hand-crafted features are commonly extracted using Fourier coefficients from sensor outputs [1], and Mel-Frequency Cepstral Coefficients (MFCCs) along with statistical descriptors of scanning speed and applied force\cite{b4}. Visual inputs were often analyzed using classical texture descriptors like Gray-Level Co-occurrence Matrix (GLCM), Gabor filters, and Local Binary Patterns (LBP)\cite{b4}. While parametric models were also explored for haptic textures, they frequently faced limitations in generalizing across diverse textures and interaction conditions. These earlier approaches, though foundational,often struggled to capture complex nonlinear relationships and exhibited less robustness compared to the subsequent deep learning-based methods\cite{b4}.

More recently, transformer-based architectures have been adopted for modeling visual-tactile perception due to their ability to capture long-range dependencies and perform effective multi-modal fusion. These models use self-attention mechanisms to align and integrate visual and tactile features across spatial and temporal domains. For instance, Chen et al., introduced Visuo-Tactile Transformers (VTTs)\cite{b5}, which extend Visual Transformers to jointly process vision and tactile feedback using self and cross-modal attention. VTT builds latent heatmap representations focusing on important task features and improves robotic manipulation tasks by leveraging complementary visuo-tactile information\cite{b5}. Researchers also introduced ViTacFormer, a vision-tactile transformer that jointly models the interaction between modalities for improved generalization under occlusion and noise\cite{b6}. These transformer-based models mark a significant advancement over earlier approaches, offering stronger robustness, better data efficiency, and improved generalization across unseen materials and conditions.

Despite promising progress, several key limitations persist in the current literature. While transformers excel at fusing modalities, there are issues when transferring visual pre-trained models (e.g., ViTs) directly to tactile data, due to domain gaps and representational differences, which can degrade performance compared
to more tactile-specialized backbones like ResNet variants\cite{b7}. Effective multi-modal fusion is complex, with some fusion mechanisms being sensitive to noise or occlusion, and requiring careful design to integrate complementary tactile and visual cues optimally \cite{b7,b8}. Additionally, feeding the vision and tactile images
directly into a deep learning model is both costly and slow. Surformer V1 overcomes these limitations by introducing handcrafted features for tactile images and PCA reduction for vision images. Implementation of an effective multi-modal approach is quite difficult with touch and go dataset\cite{b9}, as there are different
type of images: vision (wide angle) and tactile (close up Gel-sight images). Despite having a lower inference time compared to multi-modal CNN, Surformer V1 was not tailored for touch and go dataset as it leverages self and cross attention with mid level fusion, which leads to attending to irrelevant features, making it less
stable. Multi model CNN came with its own limitations including its latency and efficiency. Surformer V2 address these limitations, classify each modality separately before and after fusion. Key Contributions of this work include:

\begin{itemize}
    \item Surformer v2 significantly reduces inference time compared to previous Surformer v1 model, making it suitable for real-time applications.
    \item The model is tailored for heterogeneous input types such as first person view (wide angle) and tactile(close up texture map). Processing each modality independently and fusing decisions at a late stage improves robustness, especially when one modality is noisy or unavailable.
    \item Unlike Surformer v1, which relied on pre-extracted features, Surformer v2 performs feature extraction within the model itself.
    \item Surformer v2 accounts for the complete inference pipeline, including feature extraction time, offering a more practical assessment of real-world deployment performance.
\end{itemize}


The rest of this paper is organized as follows. Section 2 presents the Methodology, starting with an overview of the Surformer v2 architecture and detailing the Vision Branch, Tactile Branch, Weighted Fusion Layer, and Final Output Layer. Section 3 reports the Experimental Results and Discussion, covering dataset description, preprocessing, training strategies, performance evaluation, and comparisons with other multimodal approaches. Section 4 concludes the paper by summarizing key findings, discussing implications for real-time robotic perception, and suggesting future research directions.

\section{Methodology}

Surformer v2 builds upon the foundational design of Surformer v1, extending its architecture to enhance multi-modal surface classification through a more scalable and modular approach (Fig. 1). Surformer v1 relied on pre-extracted features before inputting them into the model. However, given the nature of the dataset, combining
first-person vision data with close-up tactile images from GelSight, it is counterintuitive to directly infer meaningful correlations between such distinct modalities. A more robust classification approach processes each modality independently, which is especially advantageous in scenarios where one modality is missing or unreliable (e.g., due to sensor faults). In contrast to the modality-specific dense encoders and attention-based fusion mechanism in v1, Surformer v2 introduces a late fusion strategy that classifies each modality
independently before integrating their respective outputs (logits) via a learnable weighted summation. The Vision stream in Surformer v2 employs a lightweight CNN-based classifier (Efficient V-Net), while the Tactile stream uses an encoder-only Transformer model tailored to capture spatial dependencies within tactile images. This architectural evolution enables Surformer v2 to generalize better to raw input data, simplifies training, and improves flexibility in handling diverse sensory modalities without handcrafted features or
explicit attention fusion mechanisms.

\begin{figure*}[htbp]
\centerline{\includegraphics[width=0.8\textwidth]{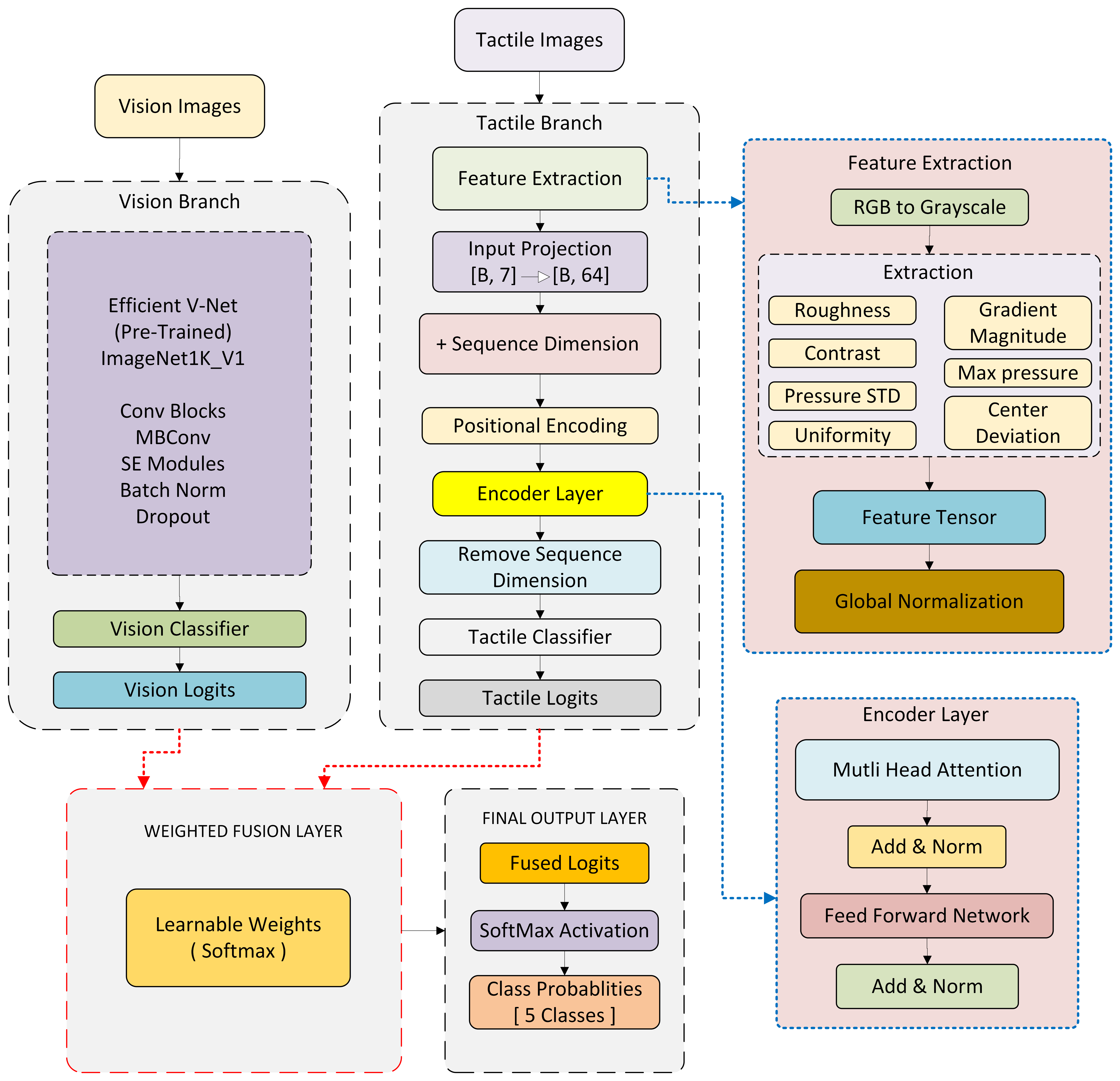}}
\caption{Architecture of Surformer v2 for multi-modal surface classification from tactile and visual data. The model accepts two input modalities: (1) vision Images, and (2) tactile images. The Surformer v2 architecture begins with two parallel branches: Vision and Tactile. The Vision branch utilizes a standard convolutional neural network (CNN) classifier called Efficient V-Net. The Tactile branch employs an encoder only transformer architecture to classify images by extracting features within the model. The hand crafted features are extracted directly in the tactile branch before passing into the transformer. Both vision and tactile images are passed through their respective branches and independently produce class logits. The outputs (logits) from both branches are combined using a learnable weighted sum. The weights are optimized during training, allowing the model to adaptively balance the importance of each modality.}
\label{fig:main}
\end{figure*}

\subsection{Vision Branch}

The vision branch of SurformerV2 employs EfficientNetV2-S \cite{b10}, a state-of-the-art convolutional neural network designed for efficient image classification. EfficientNetV2-S serves as the backbone, selected for its optimal balance between accuracy and computational efficiency. The model is initialized with ImageNet pre-trained weights, providing robust feature representations learned from millions of natural images. While this architecture uses EfficientNetV2-S, the framework is flexible and can accommodate any deep learning model based on specific classification requirements. The backbone processes RGB images [B, 3, 224, 224], B represents batch, through a series of MBConv blocks (Mobile Inverted Bottleneck Convolutions) with squeeze-and-excitation attention mechanisms. These blocks progressively extract hierarchical features from low-level edges to high-level semantic patterns. The final feature vector of 1280 dimensions captures rich visual representations suitable for surface classification. The vision branch implements a selective fine-tuning approach for optimal transfer learning. The early layers (first convolutional blocks) remain frozen to preserve low-level visual features like edges, corners, and basic textures learned during ImageNet pre-training. Only the final 20 parameters are unfrozen, allowing targeted adaptation to surface-specific visual characteristics while preventing catastrophic forgetting of foundational features. Finally, the original ImageNet classifier
is replaced with a custom two-layer Multi-Layer Perceptron (MLP) that incorporates strategic regularization, consisting of Dropout(0.1), followed by a linear layer reducing dimensions from 1280 to 256, a ReLU activation, another Dropout(0.1), and a final linear layer mapping from 256 to the number of target classes.This classifier head transforms generic EfficientNetV2-S features into domain-specific surface classification probabilities (vision logits).

\subsection{Tactile Branch}

The tactile branch begins with RGB tactile images, which are passed through a feature extraction module that returns a normalized 7-dimensional handcrafted feature vector. This vector is then projected into a 64-dimensional embedding space. Positional encoding is added for architectural consistency and potential future extensions to multi-token tactile sequences. After passing through single layer transformer encoder, the output passes through a classification head or tactile classifier which includes LayerNorm and Dropout, gradually compresses the rich feature representation into class probabilities (tactile logits).

\subsubsection{Feature Extraction and Encoder Layer}
The feature extraction process begins with RGB tactile images, which are converted into grayscale using the standard weighted average: 0.299×R + 0.587×G + 0.114×B. The hand crafted features are then carefully extracted and prepared to be fed into the model. Before training, the module computes global normalization statistics by sampling up to 1000 training examples and extracting their features. This creates dataset-specific feature mean and standard deviation tensors that ensure all 7 features have similar scales, preventing any
single feature from dominating the learning process. The model then uses a single-layer transformer encoder consisting of 4 attention heads and a Feed-Forward Layer. Each attention head focuses on different aspects of the tactile data. The encoder is 64-dimensional, and the feed-forward layer has 256 dimensions (4 times the model’s dimensionality).

\subsection{Weighed Fusion Layer}

This layer implements learnable weighted fusion that automatically determines the optimal contribution of each modality during training. The weighted fusion layer combines vision and tactile logits through
learnable parameters, with the weights being softmax-normalized. The layer receives gradients from the final loss, enabling it to discover optimal modality combinations for different surface types while supporting multi-objective training through separate auxiliary losses on individual modality outputs.

\subsection{Final Output Layer}

The final output layer transforms the fused logits from the weighted fusion layer into interpretable classification results through softmax normalization, converting raw logits into probability distributions, where each sample's probabilities sum to 1.0 and all values lie within the range of [0,1]. The predicted class is determined using the maximum likelihood principle ŷ = argmax(P), selecting the surface type with the highest probability, while the confidence score is extracted as conf = max(P) to indicate prediction certainty. This layer uses CrossEntropyLoss during training, receiving gradients from both the primary fused loss and auxiliary losses from individual modalities (weighted at 0.3), enabling the model to optimize not only the final classification accuracy but also the fusion weight balance and individual modality performance.

\section{Experimental Results and Discussion}

\subsection{Data Pre-processing and Training Strategy}

We used the Touch and Go dataset \cite{b9} for all experiments, and applied the same preprocessing pipeline as described in our previous study (Surformer v1) \cite{Surformer} excluding the feature extraction for both modalities.

The models use 80-20 train-test split with stratified sampling to ensure balanced class representation across splits. The training strategy, splits, and hyperparameter settings for the Multimodal CNN and Surformer v1 follow those established in our prior work on Surformer v1.

We also implemented early fusion  version of Multi-model CNN with a single ResNet50V2 backbone. Vision and tactile images are concatenated along the channel axis to form a 224×224×6 input, which is processed by one backbone. No pretrained weights are used (weights=none) because ImageNet weights do not support 6-channel input. Following the backbone, the extracted features were pooled and passed through a dense layer with 256 units, followed by dropout and a final classification layer. Unlike late fusion methods, this approach integrates vision and tactile information at the input level, without explicit concatenation of modality-specific feature vectors (two separate feature vectors). All other hyperparameters and optimization settings were kept consistent with multi-model CNN.

SurformerV2 employs a multi-objective training strategy that simultaneously optimizes fusion performance while keeping both modalities independent. For training, the model uses a composite loss function  defined as 
$\mathcal{L}_{\text{Total}} = \mathcal{L}_{\text{main}} + 0.3 \times \mathcal{L}_{\text{vision}} + 0.3 \times \mathcal{L}_{\text{tactile}}$, where the primary cross-entropy loss on fused predictions drives overall performance, while auxiliary losses on individual modality outputs prevent the fusion layer from over-relying on a single stream, encouraging robust feature learning in both branches.

Surformer v2 implements different learning rates: vision parameters are trained with a conservative rate of $5 \times 10^{-7}$ to preserve ImageNet knowledge, tactile transformer parameters use a higher rate of $1.5 \times 10^{-4}$ for effective from-scratch learning, and fusion weights adopt the same conservative rate for stable convergence. The training process integrates a ReduceLROnPlateau scheduler with patience of 5 epochs and a reduction factor of 0.5, adapting learning rates based on validation performance. It also includes comprehensive tracking of individual modality accuracies alongside fusion performance and supports learnable fusion weight evolution, allowing the model to automatically discover optimal modality balancing throughout training. This integrated strategy ensures that Surformer v2 achieves high classification accuracy while maintaining interpretability and robustness through strong, independent uni-modal representations.

\subsection{Results}

\begin{table}[htbp]
\centering
\caption{Performance comparison of multimodal models on the Touch and Go dataset}
\vspace{-0.2cm}
\label{tab:multimodal_performance}
\renewcommand{\arraystretch}{1.2}
\setlength{\tabcolsep}{4pt}
\begin{tabular}{|p{2.8cm}|c|c|c|}
\hline
\textbf{Model} & \textbf{Precision (\%)} & \textbf{Recall (\%)} & \textbf{F1-score (\%)} \\ \hline
Surformer v1 & 99 & 99 & 99 \\ \hline
Multimodal CNN & 100 & 100 & 100 \\ \hline
Multimodal CNN (Early Fusion) & 100 & 100 & 100 \\ \hline
Surformer v2 & 97 & 97 & 97 \\ \hline
\end{tabular}
\end{table}

\begin{table}[htbp]
\centering
\caption{Computational efficiency and model complexity comparison.}
\vspace{-0.2cm}
\label{tab:comp_efficiency}
\renewcommand{\arraystretch}{1.2}
\setlength{\tabcolsep}{4pt}
\begin{tabular}{|p{2.2cm}|c|>{\centering\arraybackslash}p{1.8cm}|>{\centering\arraybackslash}p{1.7cm}|}
\hline
\textbf{Model} & \textbf{Accuracy (\%)} & \textbf{Inference Time (ms)} & \textbf{Parameters (M)} \\ \hline
Surformer v1 & 99.40 & 0.7271 & 0.673 \\ \hline
Multimodal CNN & 100 & 5.0737 & 48.311 \\ \hline
Multimodal CNN (Early Fusion) & 100 & 4.2674 & 24.100 \\ \hline
Surformer v2 & 97.40 & 0.0239 & 20.660 \\ \hline
\end{tabular}
\end{table}

\begin{figure*}[htbp]
    \centering
    \includegraphics[width=0.8\linewidth]{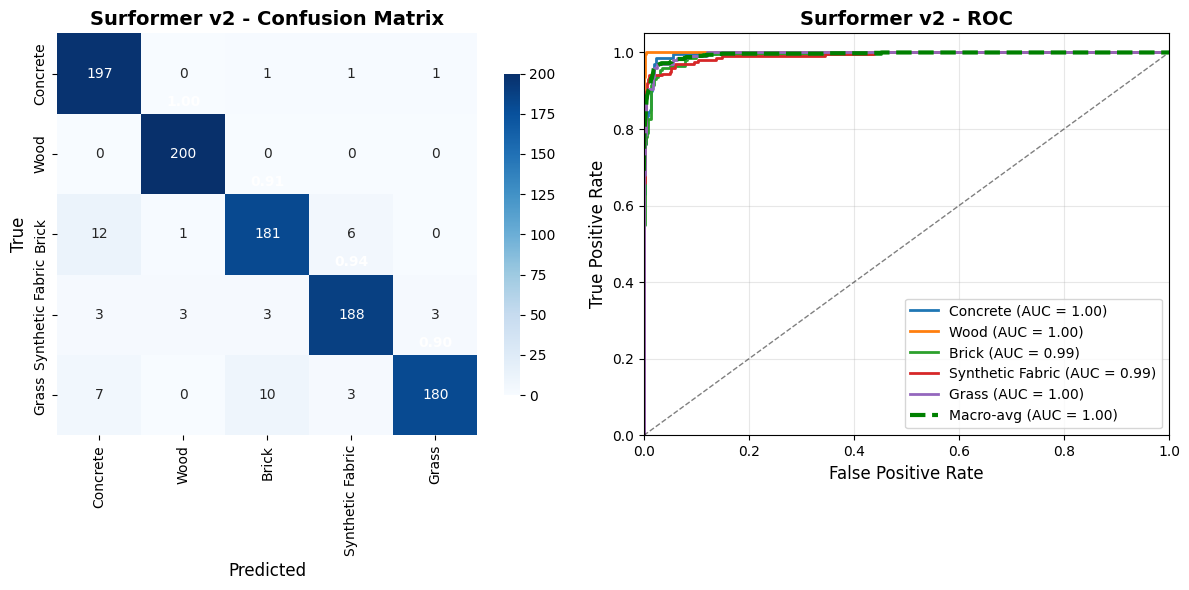}
    \caption{Performance evaluation of Surformer v2 on the Touch and Go dataset. The confusion matrix (left) illustrates the model's classification accuracy across five material classes, showing high true positive rates and minimal misclassifications. The ROC curves (right) demonstrate excellent class separability, with AUC scores of 1.00 for Concrete, Wood, and Grass, and 0.99 for Brick and Synthetic Fabric, yielding a macro-average AUC of 1.00.}
    \label{fig:tactile_features}
\end{figure*}

Table \ref{tab:multimodal_performance} presents the performance comparison of four models (Surformer v1, Multimodal CNN, Multimodal CNN with early fusion, and Surformer v2) based on precision, recall, and F1-score. Both multimodal CNN variants achieved perfect scores (100\%) across all three metrics, indicating their strong discriminative capabilities when trained on the fused vision and tactile data. Surformer v1 also performed remarkably well, achieving 99\% across all evaluation metrics, suggesting its ability to integrate multimodal signals effectively. In contrast, Surformer v2 achieved a slightly lower score of 97\%, indicating a minor drop in predictive performance. However, this trade-off must be considered in light of its significant gains in computational efficiency.

Table \ref{tab:comp_efficiency} highlights the efficiency and complexity trade-offs among the same models by reporting accuracy, inference time, and the number of trainable parameters. While the accuracy results largely mirror the earlier findings, with the CNN-based models slightly outperforming other models, Surformer v2 stands out in terms of inference efficiency and model compactness. It achieved the fastest inference time (0.0239 ms) and maintained a moderate model size (20.7M parameters), significantly outperforming both Multimodal CNN variants in computational efficiency. Notably, the standard Multimodal CNN requires over 48M parameters and exhibits the slowest inference time (5.07 ms), making it less suitable for real-time applications. Even reducing the parameters down to 28M with early fusion, the inference time didn't change significantly. Compared to Surformer v1, Surformer v2 offers a $\,\sim\,$30x speed-up in inference with only a marginal drop in accuracy (from 99.4\% to 97.4\%). These results demonstrate that Surformer v2 achieves a desirable balance between predictive performance and computational efficiency, making it a strong candidate for deployment in resource-constrained or latency-sensitive settings.

Fig. \ref{fig:tactile_features} presents the classification performance of Surformer v2 on the Touch and Go dataset. The confusion matrix (left) shows the model's ability to correctly classify instances across five material classes: Concrete, Wood, Brick, Synthetic Fabric, and Grass. The matrix reveals high diagonal values, indicating strong class-specific accuracy. Most misclassifications occur between classes with potentially overlapping tactile or visual signatures, such as Brick and Synthetic Fabric, but their frequency is minimal. For example, out of 200 true samples for each class, Surformer v2 achieves perfect classification in several cases (e.g., Wood) and maintains low misclassification counts elsewhere (e.g., 12 misclassified Brick samples as Concrete).

The ROC curve (right) further supports the model's effectiveness by illustrating its discriminative capability for each class. Area Under the Curve (AUC) values are near perfect: Concrete, Wood, and Grass each achieve an AUC of 1.00, while Brick and Synthetic Fabric achieve 0.99, indicating excellent separability. The macro-average AUC also reaches 1.00, underscoring the balanced performance of Surformer v2 across all classes. Together, these results confirm that Surformer v2 not only performs well in terms of accuracy but also exhibits strong robustness and generalization across multiple material categories.

\section{Conclusion}

In this work, we introduced Surformer v2, a decision-level multimodal framework for surface material classification that integrates visual and tactile inputs through independent modality-specific processing and a learnable weighted fusion mechanism. Compared to the previous Surformer v1 framework, which relied on pre-extracted features and mid-level cross-attention fusion, Surformer v2 embeds feature extraction directly into the architecture. This design choice improves modularity, allows each modality to specialize in its own domain, and accounts for the complete inference pipeline, including feature extraction time, yielding a $\,\sim\,$30x speed-up in inference over Surformer v1 with only a marginal drop in accuracy. Our experiments on the Touch and Go dataset demonstrate that Surformer v2 achieves competitive classification accuracy while maintaining efficiency suitable for real-time deployment in robotic systems. These results confirm the benefits of late fusion in scenarios where input modalities differ substantially in spatial scale, viewpoint, and semantic content.


\end{document}